\documentclass[10pt,twocolumn,letterpaper]{article}

\usepackage[pagenumbers]{cvpr}      

\usepackage[dvipsnames]{xcolor}
\usepackage{multirow}
\usepackage{graphicx}
\usepackage{multicol}
\usepackage{arydshln}
\usepackage{amsmath}
\usepackage{booktabs}
\usepackage{bbding}
\usepackage{textcomp}
\usepackage{tablefootnote}
\usepackage{threeparttable}
\usepackage{pifont}
\usepackage[accsupp]{axessibility}
\usepackage[accsupp]{axessibility}

\newcommand{\circled}[1]{\textcircled{\raisebox{-0.8pt}{#1}}}

\definecolor{cvprblue}{rgb}{0.21,0.49,0.74}
\usepackage[pagebackref,breaklinks,colorlinks,citecolor=cvprblue]{hyperref}

\def\paperID{5379} 
\def\confName{CVPR}
\def\confYear{2024}

\title{Contrastive Pre-Training with Multi-View Fusion for No-Reference Point Cloud Quality Assessment}

\author{Ziyu Shan$^{\rm 1}$, Yujie Zhang$^{\rm 1}$, Qi Yang$^{\rm 2}$, Haichen Yang$^{\rm 1}$ Yiling Xu$^{\rm 1}$\thanks{Corresponding author. This paper is supported in part by National Natural Science Foundation of China (62371290, U20A20185) and 111 project (BP0719010). The corresponding author is Yiling Xu(e-mail: yl.xu@sjtu.edu.cn).} \\ Jenq-Neng Hwang $^{\rm 3}$ Xiaozhong Xu$^{\rm 2}$ Shan Liu$^{\rm 2}$ \\ 
$^1$Shanghai Jiao Tong University, $^{\rm 2}$Tencent, $^{\rm 3}$ University of Washington\\ $^{\rm 1}$\{shanziyu, yujie19981026, yanghaichen, yl.xu\}@sjtu.edu.cn, \\$^{\rm 2}$\{chinoyang, shanl\}@tencent.com,$^{\rm 3}$ hwang@uw.edu
}

\begin{document}
\maketitle
\begin{abstract}
No-reference point cloud quality assessment (NR-PCQA) aims to automatically evaluate the perceptual quality of distorted point clouds without available reference, which have achieved tremendous improvements due to the utilization of deep neural networks. However, learning-based NR-PCQA methods suffer from the scarcity of labeled data and usually perform suboptimally in terms of generalization. To solve the problem, we propose a novel contrastive pre-training framework tailored for PCQA (CoPA), which enables the pre-trained model to learn quality-aware representations from unlabeled data. To obtain anchors in the representation space, we project point clouds with different distortions into images and randomly mix their local patches to form mixed images with multiple distortions. Utilizing the generated anchors, we constrain the pre-training process via a quality-aware contrastive loss following the philosophy that perceptual quality is closely related to both content and distortion. Furthermore, in the model fine-tuning stage, we propose a semantic-guided multi-view fusion module to effectively integrate the features of projected images from multiple perspectives. Extensive experiments show that our method outperforms the state-of-the-art PCQA methods on popular benchmarks. Further investigations demonstrate that CoPA can also benefit existing learning-based PCQA models.
\end{abstract}    
\section{Introduction}
\label{sec:intro}

Point clouds have emerged as a prominent 3D multimedia representation in diverse scenarios, such as autonomous driving, digital museum, and immersive gaming \cite{qi2017pointnet,afham2022crosspoint,zhang2022pointclip}. These extensive applications stem from the rich information provided by point clouds (\textit{e.g.}, geometric coordinates, color, and opacity). Nevertheless, point clouds undergo various distortions at any stage of their operation cycle (\textit{e.g.}, acquisition, compression, and transmission) before being delivered to the terminals, leading to inevitable quality degradation and perceptual loss. To optimize the quality of experience in practical applications, point cloud quality assessment (PCQA) has become one of the most fundamental and challenging problems in both industry and academic area.

\begin{figure}[t]
    \centering
\includegraphics[width=0.44\textwidth]{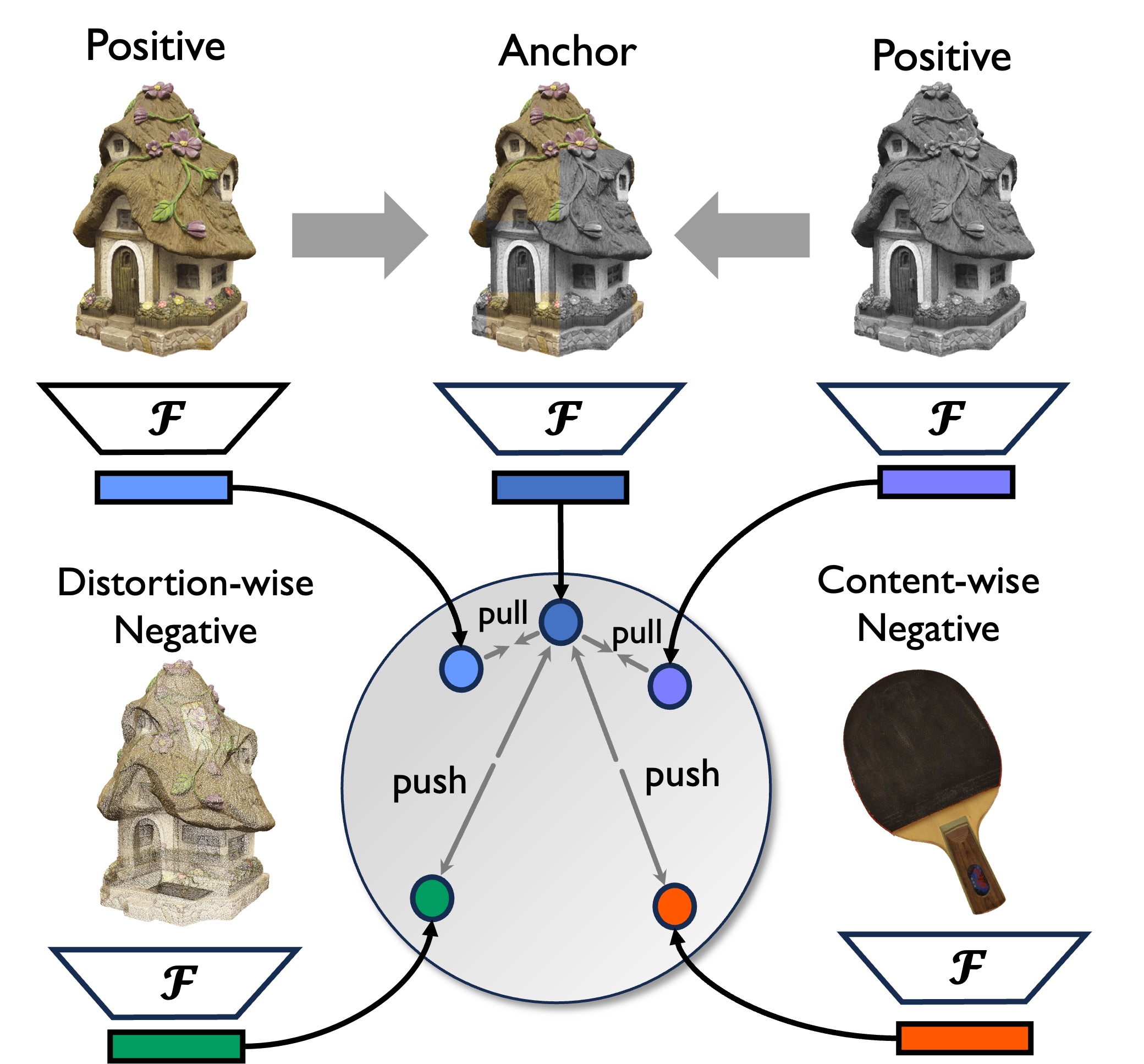}
    \caption{Illustration of our contrastive pre-training framework (CoPA). CoPA first generates anchor by randomly patch-mixing the projected images from a point cloud with different distortions, and then pre-trains the model by pulling positive samples closer to the anchor in the representation space, while pushing distortion-wise and content-wise negative samples apart.}
    \label{fig:teaser}
    \vspace{-0.5cm}
\end{figure}

PCQA methods can be classified into full-reference (FR), reduced-reference (RR) and no-reference (NR) methods, hinging on the availability of high-quality reference point clouds. In this paper, we focus on NR-PCQA considering that in most cases pristine reference point clouds are not available \cite{zhou2022blind}. 

Thanks to the tremendous improvements of deep learning networks, learning-based NR-PCQA methods have presented remarkable performance. However, the NR-PCQA problem is far from completely resolved because these data-driven methods suffer from the scarcity of labeled data.
Most PCQA datasets \cite{yang2020predicting,liu2022perceptual,liu2023point} only provide hundreds of samples with labels (\textit{i.e.,} mean opinion score (MOS)) due to the high cost of the annotation process, indicating that existing PCQA datasets are too small to train a model with good generalizability. Consequently, it limits the model performance with respect to cross-dataset evaluation. 

Researchers have proposed several methods to address this problem. An intriguing way \cite{yang2022no} is to take advantage of the rich subjective scores of natural images and infer point cloud quality through unsupervised adversarial domain adaptation. Nevertheless, the 2D-to-3D perceptual adaptation is considerably difficult due to significant discrepancies of content and distortion characteristics between natural images and point clouds.  As a result, the learned universal encoder fails to extract adequately effective features to infer point cloud quality. Another common strategy \cite{liu2021pqa, shan2023gpa} is to employ distortion-related auxiliary tasks (\textit{e.g.}, distortion type prediction) to equip models with the preliminary capability of recognizing distortion patterns. However, the models trained by these auxiliary tasks are not robust when adapted to new datasets with unseen distortion types.

Contrastive learning is another potential choice to promote the generalizability of NR-PCQA models because of its ability to leverage a large amount of unlabeled data. In the classical contrastive learning paradigm, for a given anchor target (often obtained through data augmentation), the model intends to learn representations by pulling similar samples (\textit{i.e.}, positive samples) closer and pushing dissimilar ones (\textit{i.e.}, negative samples) apart in the representation space. After this pre-training, the model is fine-tuned for different downstream tasks. However, commonly used anchor generation methods (\textit{e.g.}, converting to grayscale images in image classification tasks) can introduce extra distortions and degrade the original perceptual quality, thus hindering the quality-aware representation learning in PCQA task. Furthermore, pre-trained models under the classical contrastive learning paradigm mainly focus on high-level semantic information \cite{he2020momentum,chen2020simple}, while perceptual quality is determined by both high-level content and low-level distortion.


To solve these problems, we propose a novel 
\textbf{Co}ntrastive pre-training framework tailored for \textbf{P}CQ\textbf{A} (CoPA). 
CoPA has two main steps following the contrastive learning paradigm: 1) Anchor generation by patch mixing. CoPA applies random rotations to the point clouds to take advantage of the rich quality information of point clouds from multiple perspectives, and then projects the point clouds impaired by different distortions into images. Subsequently, CoPA randomly mixes the local patches of the projected images to form a mixed image with multiple distortions, which is used as an anchor in the representation space, as illustrated in \cref{fig:teaser}. The anchor generation process does not introduce additional distortions, thus preserving intrinsic quality information within the projected images. Moreover, due to the random rotation and patch mixing, CoPA can generate an unbounded number of anchors to form an extensive set of training pairs, allowing for a comprehensive pre-training of the network. 
2) Content-wise and distortion-wise contrast. Inspired by the observation that perceptual quality is correlated with both content and distortion pattern \cite{chen2021contrastive,zhao2023quality,saha2023re}: for an anchor image, CoPA treats the projected images that form the anchor, as positive samples; images projected from different reference point clouds (content-wise) and images from the same reference point cloud but with unrelated distortions (distortion-wise) are regarded as negative samples. Following the contrastive learning paradigm, we finally obtain a pre-trained encoder, $\mathcal{F}$, that is expected to extract quality-aware features from projected images. The encoder $\mathcal{F}$ is then incorporated into our model and fine-tuned with labeled data. 

In fine-tuning stage, point clouds are first projected into multi-view images from different perspectives to mimic the subjective observation process. Then, a semantic-guided multi-view fusion module is proposed to integrate the multi-view quality-aware features generated by $\mathcal{F}$. Concretely, the multi-view images are stitched into a composed image, which is fed into a 2D backbone $\mathcal{G}$ pre-trained on ImageNet \cite{deng2009imagenet} to extract the high-level semantic feature. Subsequently, the semantic feature guides the fusion of multi-view features through a cross-attention mechanism to obtain the final feature, followed by a quality score regression module to predict objective scores. We summarize the main contributions as follows:

 \begin{itemize}
     \item To tackle the challenge of label scarcity, we propose a contrastive pre-training framework (CoPA) tailored for PCQA. By generating anchor samples through local patch mixing and carefully designing positive/negative samples, CoPA enables the model to learn quality-aware representations to boost NR-PCQA performance. 
     \item We propose a semantic-guided multi-view fusion module to effectively integrate the quality-aware features of the projected images from different perspectives in the fine-tuning stage.
     \item Extensive experiments show that the proposed method presents superior overall performance and generalization ability compared to the state-of-the-art NR-PCQA methods. Further investigation shows that CoPA can be integrated into other projection-based NR-PCQA methods with noticeable gains.
 \end{itemize}
\section{Related Works}
\label{sec:related}
\subsection{No-Reference Point Cloud Quality Assessment}

NR-PCQA aims to evaluate the perceptual quality of distorted point clouds without available references. {Early NR-PCQA metrics typically train a regressor to
obtain quality scores based on hand-crafted features.} Zhang et al. \cite{zhang2022no} propose an NR-PCQA metric that projects point clouds into feature domains based on geometry and color and obtains the predicted MOS using a support vector machine (SVM). Zhou et al. \cite{zhou2022blind} develop a general and efficient metric based on structure-guided resampling. {Handcrafted features have explicit meanings and have shown decent evaluating ability. However, they are usually designed based on the still limited understanding of point cloud distortions, which are not comprehensive and thus are relatively limited in complex distortion environments.}

{With the boom of deep learning, many learning-based methods have been proposed for NR-PCQA.} ResSCNN \cite{liu2023point} employs a voxel-based sparse 3D-CNN for quality prediction. Fan et al. \cite{fan2022no} transform point clouds into video sequences to infer the visual quality. PM-BVQA \cite{tao2021point} uses an element-wise summation method to fuse the cross-modal and multi-scale features. Tliba et al. \cite{tliba2022representation} propose a shallow model for interpolating compression-induced quality scores. PCQA-Graphpoint \cite{tliba2023pcqa} employs a two-stream architecture to process geometry and color distortion, and combine features that leverage the attention mechanism. MVAT \cite{mu2023multi} aggregates the content and positional context via a multi-view aggregation transformer framework. $\psi$-Net \cite{xiong2023psi} extracts geometric and color structural information from 3D patches by mapping the position vectors of neighboring points to weights. MM-PCQA \cite{zhang2022mm} utilizes a novel multi-modal learning approach to assess the perceptual quality. However, these methods are all data-driven and thus suffer from the label paucity problem in PCQA.

{To reduce dependence on labeled data}, IT-PCQA \cite{yang2022no} leverages the rich subjective scores of natural images and try to evaluate point cloud quality through unsupervised domain adaptation. Nevertheless, the domain gap hinders optimal training and leads to unsatisfactory performance. PQA-Net \cite{liu2021pqa} and GPA-Net \cite{shan2023gpa} both utilize distortion type prediction as an auxiliary task to integrate distortion-related knowledge into models. However, these methods ignore the correlation between distortions and present inferior performance for unseen distortion types.

\subsection{Contrastive Learning}
Contrastive learning aims to learn meaningful representations by maximizing the similarity between similar data samples and minimizing it between dissimilar ones \cite{oord2018representation,ye2023fedfm}. MoCo \cite{he2020momentum} uses a momentum update mechanism to maintain a long queue of negative samples for contrastive learning. SimCLR \cite{chen2020simple} utilizes negative samples with a much larger batch size and constructs a rich family of data augmentations. However, most pre-trained models using these contrastive learning paradigms mainly focus on high-level semantic information, ignoring critical low-level distortions.

In image/video quality assessment scenarios, CONTRIQUE \cite{madhusudana2022image} utilizes a simple framework with quality-preserving augmentations to learn generalizable representations that perform well on synthetic and realistic distortions. Re-IQA \cite{saha2023re} develops a holistic approach to assess the quality of images in the wild by leveraging the complementary content and image quality-aware features. QPT \cite{zhao2023quality} supposes that the quality of patches from a distorted image should be similar, but vary from patches from different images and the same image with different degradations. CSPT \cite{chen2021contrastive} learns useful feature representation by using distorted video samples not only to formulate content-aware distorted instance contrasting but also to constitute an extra self-supervision signal for the distortion prediction task. However, when focusing on PCQA task, the significant volume of data brings challenges to the contrastive learning paradigm. To tackle this problem, we propose a projection-based contrastive pre-training framework considering the fact that point clouds are perceived from 2D perspective via projecting them on the screen in the subjective experiments of popular PCQA datasets \cite{yang2020predicting,liu2023point,liu2022perceptual}. Furthermore, after model pre-training, we fine-tune the model using a proposed multi-view fusion module to integrate the features from different perspectives.
\section{Method}
\begin{figure*}[h]
    \centering
    \vspace{-0.3cm}
    \includegraphics[width=0.9\textwidth]{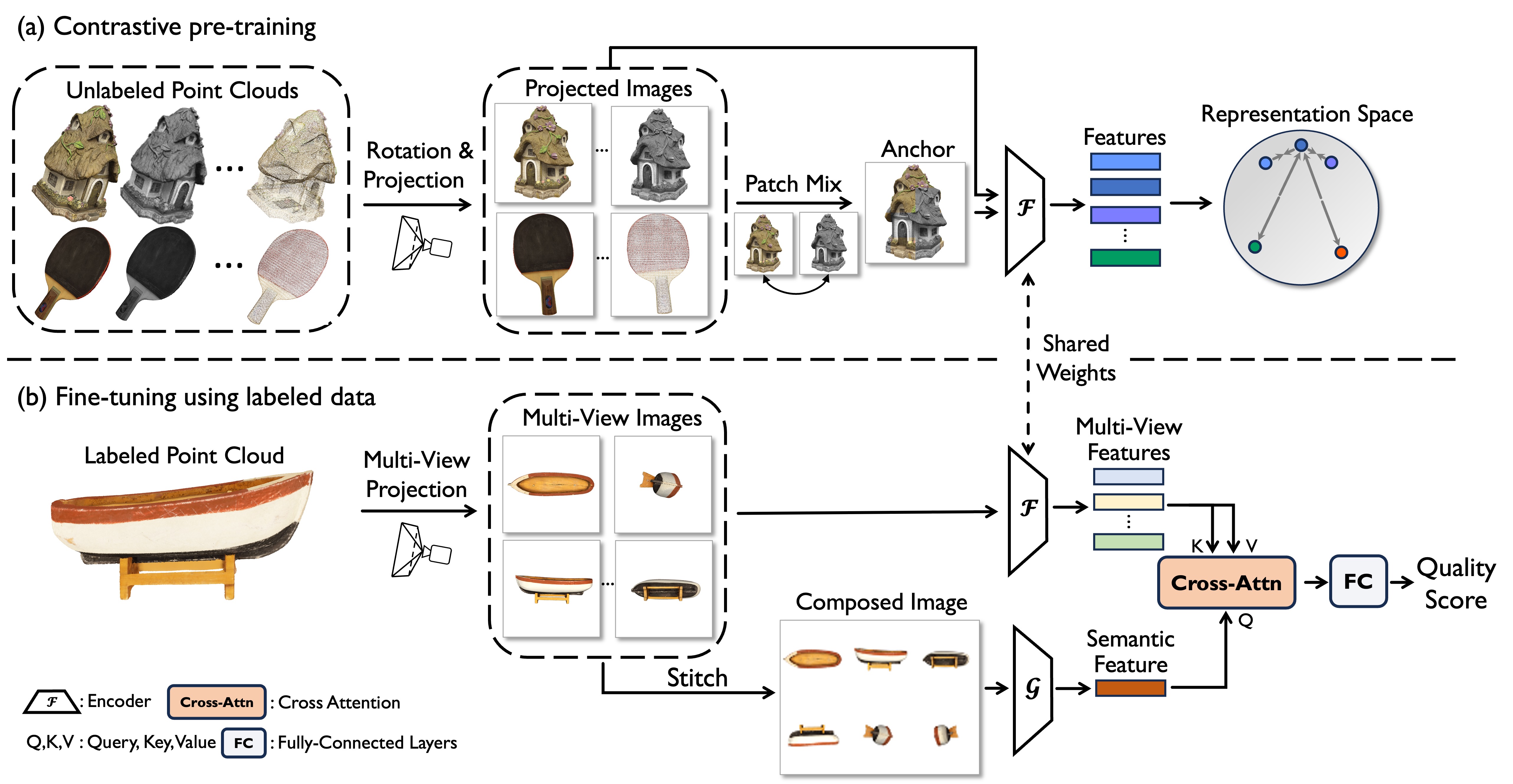}
    \caption{\textbf{Framework of the proposed method.} The framework mainly consists of two stages: (a) Contrastive pre-training. Unlabeled point clouds are projected into single-view images, and anchors are generated by local patch mixing. A single-view image encoder $\mathcal{F}$ is pre-trained by pulling the defined positive samples to the anchors in the representation space and pushing the negative samples apart.  (b) Fine-tuning with labeled data. The labeled point cloud is projected into multi-view images encoded by the pre-trained encoder $\mathcal{F}$. Then, the multi-view images are stitched to extract semantic features through a 2D backbone $\mathcal{G}$, which guides the fusion of multi-view images using the cross-attention mechanism. Finally, the quality score is regressed by the fully-connected layers.}
    \label{fig:overview}
\end{figure*}
\subsection{Overview}
The goal of the proposed method is to pre-train an encoder that can effectively extract quality-aware features and then fine-tune it with labeled data to accurately predict the quality scores. Therefore, the framework of our model is divided into two stages: 1) In the first stage, we pre-train a quality-aware encoder $\mathcal{F}$ using CoPA: each point cloud is first randomly rotated and projected into images. Then, we generate the anchor by local patch mixing and feed it into encoder $\mathcal{F}$, facilitating $\mathcal{F}$ to learn quality-aware representations through both content-wise and distortion-wise contrast. 2) In the second stage, $\mathcal{F}$ is fine-tuned with labeled data to predict quality scores: the labeled point clouds are first projected into multi-view images and then encoded by $\mathcal{F}$. Afterward, the multi-view features are fused to regress objective scores through a semantic-guided cross-attention operation, where the global semantic feature used to guide fusion is extracted by another 2D backbone $\mathcal{G}$ pre-trained on ImageNet \cite{deng2009imagenet} for image classification. 

\subsection{Contrastive Pre-Training}
\label{sec:contrastive}
Given $N$ reference point clouds $p^{(1)}$ $,\cdots, p^{\left( N \right)} $ with varied contents, each point cloud $p^{\left( n \right)}$ is degraded with $D$ types of distortion to form a distorted group $\{p^{\left( n \right)}_d |_{d=1}^D\}$ that shares consistent content but different distortions. 

\noindent\textbf{Point Cloud Projection.} Instead of performing contrastive learning directly in 3D space, we project point clouds into 2D images for two main advantages: 1) Pixel-to-pixel correspondence is established between each two projected images sampled from $\{p^{\left( n \right)}_d |_{d=1}^D\}$, which facilitates the subsequent anchor generation. 2) Projected images have fixed and smaller data sizes compared to point clouds, which not only benefits parallel processing, but also increases batch size --- a critical factor for contrastive learning \cite{chen2020simple}.

To perform the projection, given a distorted point cloud $p^{\left( n \right)}_d$, we first apply random rotations $\mathcal{R} $ to capture the visual information from different perspectives. For each rotation, the rotation degree is shared among all samples in $\{p^{\left( n \right)}_d |_{d=1}^D\}$ to ensure the content consistency. Then we apply a geometric normalization operation $\mathcal{N}$ to translate the rotated point cloud $\mathcal{R}(p^{\left( n \right)}_d)$ to the origin and rescale it to the unit ball to achieve a consistent spatial scale. Finally, we render $\mathcal{N}(\mathcal{R}( p^{\left( n \right)}_d))$ into an image $x^{\left( n \right)}_d \in \mathbb R^{H\times W \times C}$ under a fixed viewing distance.
Different rotation angles are applied for each point cloud to comprehensively capture the visual information from multiple viewpoints. 

\noindent\textbf{Anchor Generation.} Different from commonly used data augmentation methods \cite{chen2021contrastive, zhao2023quality} (e.g., cropping, noise addition) that can bring unexpected distortions to affect the original quality information, we generate anchors by mixing local image patches to better preserve local distortion patterns, as illustrated in \cref{fig:teaser}.

For two distorted projected images $x^{\left( n \right)}_{d_1}$ and $x^{\left( n \right)}_{d_2}$ that share the same reference content but different distortions, we first partition $x^{\left( n \right)}_{d_1}$ and $x^{\left( n \right)}_{d_2}$ into non-overlapping $16\times16$ patches following \cite{dosovitskiy2020image}, and mix the patches using a random binary mask $\mathbf{M} \in \{0,1\}^{H\times W}$. We define the mixing operation as follows:
\begin{equation}
    \widetilde x^{\left( n \right)}_{d_{1,2}} = \mathbf{M} \odot  x^{\left( n \right)}_{d_1} + (\mathbf{1}-\mathbf{M}) \odot   x^{\left( n \right)}_{d_2},
\end{equation}
where $\mathbf{M}$ consists of $16\times16$ binary blocks, and the elements share the same value (0 or 1) within each binary block. $\mathbf{1}$ is a binary mask filled with ones and $\odot$ denotes element-wise multiplication. $\mathbf{M}$ is randomly generated so that numerous anchors and training pairs can be created.

After local patch mixing, the anchor $x^{\left( n \right)}_{d_{1,2}}$ contains two different distortions $d_1$ and $d_2$ but shares the same reference content with $x^{\left( n \right)}_{d_1}$ and $x^{\left( n \right)}_{d_2}$. Therefore, we can assume that the anchor $x^{\left( n \right)}_{d_{1,2}}$ is ``similar" to $x^{\left( n \right)}_{d_1}$ and $x^{\left( n \right)}_{d_2}$, which will be considered positive samples in the contrastive pre-training, as detailed in the following paragraph. 

\noindent \textbf{Content-wise and Distortion-wise Contrast.} We enable the pre-trained model to be attentive to both high-level content and low-level distortion via carefully designed positive/negative samples. Given the anchor $\widetilde x^{\left( n \right)}_{d_{1,2}}$, $x^{\left( n \right)}_{d_1}$ and $x^{\left( n \right)}_{d_2}$ are considered positive samples due to the consistent content and the correlated distortion, while other samples are treated as negative ones that can be further divided into two parts. First, the samples of the set $\{ x^{\left( n \right)}_d |_{d\neq d_1,d_2}^D\}$ are denoted as distortion-wise negative samples due to the same reference content but different distortions. Second, distorted images $\{ x^{\left( m \right)}_d |_{m\neq n,d=1}^{N,D} \}$ originating from other reference point clouds (\textit{i.e.}, $p^{(m)}$) are denoted as content-wise negative samples. Notably, there may exist some corner samples that have different contents but present similar qualities, but they are negligible considering the small proportion as verified in \cite{zhao2023quality}.

In the pre-training process, given an input sample $ x^{\left( n \right)}_d$ and the encoder $\mathcal{F}$, we denote the encoded feature as $f^{\left( n \right)}_d = \mathcal{F}(x^{\left( n \right)}_d)/||\mathcal{F}(x^{\left( n \right)}_d)||$, after a $\mathcal{L_2}$ normalization. Then the distortion-wise and content-wise contrastive pre-training loss $\mathcal{L}_{d}, \mathcal{L}_{c}$ for the anchor feature $f^{\left ( n \right )}_{d_{1,2}}$ and the corresponding positive/negative samples can be formulated as:

\begin{equation}
\begin{split}
   \mathcal{L}_{d} = &-r\log \frac{\exp (f^{\left ( n \right )}_{d_{1,2}} \cdot f^{\left ( n \right )}_{d_1}/\tau) }{\sum_{d \neq d_1, d_2}^D \exp(f^{\left ( n \right )}_{d_{1,2}} \cdot f^{\left( n \right)}_d / \tau)} \\
   & -(1-r)\log \frac{\exp (f^{\left ( n \right )}_{d_{1,2}} \cdot f^{\left ( n \right )}_{d_2} / \tau) }
    {\sum_{d \neq d_1, d_2}^D \exp(f^{\left ( n \right )}_{d_{1,2}} \cdot f^{\left( n \right)}_d/ \tau)},
    \end{split}
    \end{equation}
    \begin{equation}
    \begin{split}
    \mathcal{L}_{c} = &-r\log \frac{\exp (f^{\left ( n \right )}_{d_{1,2}} \cdot f^{\left ( n \right )}_{d_1}/\tau) }{\sum_{m\neq n}^N \sum_{d=1}^D \exp(f^{\left ( n \right )}_{d_{1,2}} \cdot f^{\left( m \right)}_d / \tau)} \\
    & -(1-r) \log \frac{\exp (f^{\left ( n \right )}_{d_{1,2}} \cdot f^{\left ( n \right )}_{d_2}/\tau)}{\sum_{m\neq n}^N \sum_{d=1}^D \exp(f^{\left ( n \right )}_{d_{1,2}} \cdot f^{\left( m \right)}_d / \tau)}
    \end{split}
    \label{eq:pre_loss}
\end{equation}
where $(\cdot )$ denotes dot-product, $\tau$ is a temperature hyper-parameter. And $r = \sum\limits_{h=1}^H\sum\limits_{w=1}^W \mathbf{M}(h,w) / HW$ represents the masking ratio. 
Intuitively, the more patches of the anchor come from $x^{\left( n \right)}_{d_1}$ (or $x^{\left( n \right)}_{d_2}$), the closer $x^{\left( n \right)}_{d_1}$ (or $x^{\left( n \right)}_{d_2}$) should be pulled toward the anchor in the representation space.

Overall, the contrastive pre-training loss function can be formulated as:
\begin{equation}
    \mathcal{L}_{pre} = \lambda \mathcal{L}_{d} + (1-\lambda)\mathcal{L}_{c}
\end{equation}
where $\lambda$ is the weighting coefficient .

Considering the large scale of content-wise negative samples, it is non-trivial to extract quality-aware features from samples with all contents at each iteration. Therefore, we adopt the momentum contrast strategy \cite{he2020momentum} to reduce computational consumption, where a dynamic queue is established to enlarge the available size for contrasting. Meanwhile, the encoder parameter $\theta_{\mathcal{F}}$ is updated smoothly using the weighted sum of its previous weights. Once the update of the encoder parameter has converged, the pre-trained single-view encoder $\mathcal{F}$ can be used as a backbone in downstream tasks to extract quality-aware features.

\subsection{Fine-Tuning with Labeled Data}
After obtaining the pre-trained encoder $\mathcal{F}$, we next fine-tune it using labeled point clouds. {To mimic the multi-view characteristic when observing 3D object, we first project each point cloud into images from different viewpoints and encode these projected images using $\mathcal{F}$.}
Then, a multi-view fusion module is proposed to take advantage of global semantic information to integrate the multi-view features, considering that different views do not contribute equally to quality decision. 

\noindent\textbf{Multi-View Projection.} For a point cloud $p$ with label $q$, we normalize it with $\mathcal{N}(\cdot)$ and render $\mathcal{N}( p)$ into multi-view images $\{  x_i \in \mathbb R^{H\times W \times C} |_{i=1}^6 \}$ from six perpendicular viewpoints (\textit{i.e.,} along the positive and negative directions of x,y,z axes) with fixed viewing distance. The rendering configuration is identical to \cref{sec:contrastive}.

\noindent\textbf{Semantic-Guided Multi-View Fusion.} To fuse the multi-view features derived from six projected images, another 2D encoder $\mathcal{G}$ pre-trained on ImageNet \cite{deng2009imagenet} (image classification task) is employed to extract global semantic feature. Following the practices in \cite{yang2022no}, we first stitch the multi-view images into a composed image ${x}_c \in \mathbb R^{2H \times 3W \times C}$, and next encode it with $\mathcal{G}$ to obtain the semantic feature $g$. Then we utilize $ g$ to guide the fusion of $\{ f_i|_{i=1}^6 \}$ through a simple multi-headed cross-attention mechanism $\mathcal{M}$. 

Given three common sets of inputs: query set $Q$, key set $K$, and value set $V$, we define the $\mathcal{M}$ as:
\begin{equation}
\begin{split}
   & \mathcal{M}(Q, K, V) = (\Gamma_1 \oplus \Gamma_2 \cdots \oplus \Gamma_{N_\Gamma})W,\\
   & \Gamma_{\mu} = \mathcal{A}(QW_\mu^Q, KW_\mu^K, VW_\mu^V) |_{\mu=1}^{N_\Gamma},\\
   & \mathcal{A}(Q,K,V) = \text{softmax} \left( \frac{QK^T}{\sqrt{d_f}}\right)V
\end{split}
\end{equation}
where $\Gamma_\mu$ is the $\mu$-th head, $\mathcal{A}$ denotes attention function. 
$W, W^Q, W^K, W^V$ are learnable linear mappings and $d_f$ is a scaling factor. Then the multi-view fusion can be formulated as:
\begin{equation}
     F = \mathcal{M} ({g}, \{ f_i |_{i=1}^6 \}, \{ f_i |_{i=1}^6 \})
\label{eq:fusion}
\end{equation}
where $ F$ is the fused feature. In \cref{eq:fusion}, intuitively, the semantic feature acts as a query and computes similarity with each $ f_i$ to find the semantically active viewpoints and fuse their quality-aware features attentively.

\subsection{Quality Regression and Loss Function}

After multi-view fusion, the final feature $ F$ is fed into two-fold fully-connected layers to regress the predicted quality score $\hat q$. Inspired by \cite{zhang2022mm}, our loss function includes two parts: mean square error (MSE) and rank error. The MSE optimizes the model to improve the prediction accuracy, which is formulated as:
\begin{equation}
    \mathcal{L}_{mse} = \frac{1}{B} \sum_b^B (\hat q_b - q_b)^2
\end{equation}
where $\hat q_b$ is the predicted quality score of $b$-th sample in a mini-batch with the size of $B$, and $q_b$ is the corresponding ground truth MOS.

To better recognize quality differences for the point clouds with close MOSs, we use a differential ranking loss \cite{sun2022deep} to model the ranking relationship between $\hat q$ and $q$:
\begin{equation}
\begin{aligned}
    &\mathcal{L}_{rank} = \frac{1}{B^2} \sum_{i=1}^B \sum_{j=1}^B \mathcal{L}_{rank}^{ij}, \\ 
    \mathcal{L}_{rank }^{i j}\!=\!\max\! &\left(0,\left|q_{i}-q_{j}\right|\!-\!e\left(q_{i}, q_{j}\right)\! \cdot\! \left(\hat{q}_{i}-\hat{q}_{j}\right)\right), \\
    &e\left(q_{i}, q_{j}\right)=\left\{\begin{array}{r}
1, q_{i} \geq q_{j}, \\
-1 , q_{i} < q_{j},
\end{array}\right.
\end{aligned}
\end{equation}

Then the overall loss function for fine-tuning can be calculated as the sum of MSE loss and ranking loss:
\begin{equation}
    \mathcal{L}_{fine} = \alpha \mathcal{L}_{mse} + (1-\alpha) \mathcal{L}_{rank}
\end{equation}
where the hyper-parameter $\alpha$ is to balance the two losses.

\section{Experiments}
\begin{table*}[t]\small
\centering
\renewcommand\arraystretch{0.95}
\vspace{-0.3cm}
\caption{Performance results on the LS-PCQA \cite{liu2023point}, SJTU-PCQA \cite{yang2020predicting} and WPC \cite{liu2022perceptual} databases. ``P" and ``I" stand for the method is based on the point cloud and image modality, respectively. ``$\uparrow$"/``$\downarrow$" indicates that larger/smaller is better. The best performance results are marked in {\bf\textcolor{red}{RED}} and the second results are marked in {\bf\bf\textcolor{blue}{BLUE}} for both FR-PCQA and NR-PCQA methods. ``FT'' indicates fine-tuning.}
\begin{tabular}{c:c:c|ccc|ccc|ccc}
\toprule
    \multirow{2}{*}{Ref}&\multirow{2}{*}{Modal}&\multirow{2}{*}{Methods} & \multicolumn{3}{c|}{LS-PCQA \cite{liu2023point}} & \multicolumn{3}{c|}{SJTU-PCQA \cite{yang2020predicting}} & \multicolumn{3}{c}{WPC \cite{liu2022perceptual}} \\ \cline{4-12}
        & & & SROCC$\uparrow$      & PLCC$\uparrow$        & RMSE $\downarrow$  & SROCC$\uparrow$      & PLCC$\uparrow$        & RMSE $\downarrow$    & SROCC$\uparrow$      & PLCC$\uparrow$          & RMSE $\downarrow$ \\ \hline
\multirow{9}{*}{FR} 
 &P&MSE-p2po  & 0.325 & 0.528 & 0.158 & 0.783 & 0.845 & 0.122 & 0.564 & 0.557 & 0.188 \\
 &P&HD-p2po   & 0.291 & 0.488 & 0.163 & 0.681 & 0.748 & 0.156 & 0.106 & 0.166 & 0.222 \\
 &P&MSE-p2pl & 0.311 & 0.498 & 0.160 & 0.703 & 0.779 & 0.149 & 0.445 & 0.491 & 0.199 \\
 &P&HD-p2pl   & 0.291 & 0.478 & 0.163 & 0.617 & 0.661 & 0.177 & 0.344 & 0.380 & 0.211 \\
 &P&PSNR-yuv  & \bf\textcolor{blue}{0.548} & \bf\textcolor{blue}{0.547} & 0.155 & 0.704 & 0.715 & 0.165 & 0.563 & 0.579 & 0.186 \\
  &P&PointSSIM & 0.180 & 0.178 & 0.183 & 0.735 & 0.747 & 0.157 & 0.453 & 0.481 & 0.200 \\
 &P&PCQM      & 0.439 & 0.510 & \bf\textcolor{blue}{0.152} & 0.864 & 0.883 & \bf\textcolor{blue}{0.112} & \bf\textcolor{red}{0.750} & \bf\textcolor{red}{0.754} & \bf\textcolor{red}{0.150} \\
 &P&GraphSIM  & 0.320 & 0.281 & 0.178 & 0.856 & 0.874 & 0.114 & 0.679 & 0.693 & 0.165 \\
&P&MS-GraphSIM & 0.389 & 0.348 & 0.174 & \bf\textcolor{blue}{0.888} & \bf\textcolor{blue}{0.914} & \bf\textcolor{red}{0.096} & \bf\textcolor{blue}{0.704} & \bf\textcolor{blue}{0.718} & \bf\textcolor{blue}{0.159} \\
 &P&MPED  & \bf\textcolor{red}{0.659} & \bf\textcolor{red}{0.671} & \bf\textcolor{red}{0.138} & \bf\textcolor{red}{0.898} & \bf\textcolor{red}{0.915} & \bf\textcolor{red}{0.096} & 0.656 & 0.670  & 0.169
 \\\hdashline
\multirow{6}{*}{NR} 
 &I&PQA-Net   & 0.588 & 0.592 & 0.202 & 0.659 & 0.687 & 0.172 & 0.547 & 0.579 & 0.189\\
 &I&IT-PCQA   & 0.326 & 0.347 & 0.224 & 0.539 & 0.629 & 0.218 & 0.422 & 0.468 & 0.221 \\
 &P&GPA-Net   & 0.592 & 0.619 & 0.186 & \bf\textcolor{blue}{0.878} & 0.886 & 0.122 & 0.758 & 0.769 & 0.162 \\
 &P&ResSCNN   & \bf\textcolor{blue}{0.594} & \bf\textcolor{blue}{0.624} & \bf\textcolor{blue}{0.172} & 0.834 & 0.863 & 0.153 & 0.735 & 0.752 & 0.177 \\
 &P+I&MM-PCQA   & 0.581 & 0.597 & 0.189 & 0.876 & \bf\textcolor{blue}{0.898} & \bf\textcolor{blue}{0.109} & \bf\textcolor{blue}{0.761} & \bf\textcolor{blue}{0.774} & \bf\textcolor{blue}{0.149} \\
 &P&\textbf{CoPA+FT} & \bf\textcolor{red}{0.613}& \bf\textcolor{red}{0.636} & \bf\textcolor{red}{0.161}  & \bf\textcolor{red}{0.897} & \bf\textcolor{red}{0.913} &
 \bf\textcolor{red}{0.092} & \bf\textcolor{red}{0.779} & \bf\textcolor{red}{0.785} &  \bf\textcolor{red}{0.144} \\
\bottomrule
\end{tabular}
\label{tab:main}
\end{table*}

\subsection{Datasets and Evaluation Metrics}

\noindent\textbf{Datasets.} Our experiments are based on three commonly used PCQA datasets, including LS-PCQA \cite{liu2023point}, SJTU-PCQA \cite{yang2020predicting}, and WPC \cite{liu2022perceptual}. The pre-training is based on LS-PCQA, which is a large-scale PCQA dataset and contains 104 pristine reference point clouds and 24,024 distorted point clouds, and each reference point cloud is impaired with 33 types of distortions (\textit{e.g.,} V-PCC, G-PCC) under 7 levels.
For a fair comparison, the model is fine-tuned on all three datasets separately using labeled data, where SJTU-PCQA includes 9 reference point clouds and 378 distorted samples impaired with 7 types of distortions (\textit{e.g.,} color noise, downsampling) under 6 levels, while WPC contains 20 reference point clouds and 740 distorted sampled disturbed by 5 types of distortions (\textit{e.g.,} compression, gaussian noise). 

\noindent\textbf{Evaluation Metrics.} Three widely adopted evaluation metrics are employed to quantify the level of agreement between predicted quality scores and MOSs: Spearman rank order correlation coefficient (SROCC), Pearson linear correlation coefficient (PLCC), and root mean square error (RMSE). To ensure consistency between the value ranges of the predicted scores and subjective values, nonlinear Logistic-4 regression is used to align their ranges.

\subsection{Implementation Details}
\label{sec:implementation}
Our experiments are performed using PyTorch \cite{paszke2019pytorch} on NVIDIA 3090 GPUs. All point clouds are rendered into projected images with a spatial resolution of $512\times512$ by PyTorch3D \cite{ravi2020accelerating}. 
Following the contrastive learning paradigm, the experiment is performed in two stages:

\noindent\textbf{Pre-Training.} We employ a Swin-T \cite{liu2021swin} as encoder $\mathcal{F}$ to be pre-trained for 200 epochs. The momentum parameter updating follows the configuration in \cite{he2020momentum}.
We use the SGD optimizer \cite{sutskever2013importance} with weight decay 0.0001, momentum of 0.95, and batch size of 128. The learning rate is 0.005 and decayed by 0.2 every 10 epochs. $\lambda$ and $\tau$ are set to 0.3 and 0.2. Each point cloud is randomly rotated 6 times before projection, and the masking ratio $r$ is bounded to $[0.25, 0.75]$ to ensure that local patches are sufficiently mixed and mitigate the influence of projected images' backgrounds.

\noindent\textbf{Fine-Tuning.} We use a ResNet50 \cite{he2016deep} pre-trained on ImageNet \cite{deng2009imagenet} as encoder $\mathcal{G}$ to extract semantic features. The encoded features from $\mathcal{F}$ and $\mathcal{G}$ are projected to a channel size of 1024 by one-layer MLPs. The multi-headed cross-attention employs 8 heads and $d_f$ is empirically set to 1024. We use the Adam optimizer \cite{kingma2014adam} with weight decay of 0.0001 and batch size of 16. The hidden dimension of fully-connected layers is set to 512. The learning rate is initialized with 0.003 and decayed by 0.9 every 5 epochs. For LS-PCQA, the model is trained for 20 epochs, while 150 epochs for SJTU-PCQA and WPC. The hyper-parameter $\alpha$ is set to 0.5.

Considering the limited dataset scale, in the fine-tuning stage, 5-fold cross validation is used for SJTU-PCQA and WPC to reduce content bias. Take SJTU-PCQA for example, in each fold, the dataset is split into train-test with ratio 7:2 according to the reference point clouds, where the performance on testing set with minimal training loss is recorded and then averaged across five folds to get the final result. A similar procedure is repeated for WPC where the train-test ratio is 4:1. As for the large-scale LS-PCQA, it is split into train-val-test with a ratio around 8:1:1 (no content overlap exists). The result on the testing set with the best validation performance is recorded. Note that the pre-training is only conducted on the training set of LS-PCQA.

\subsection{Comparison with State-of-the-art Methods}
15 state-of-the-art PCQA methods are selected for comparison, including 10 FR-PCQA and 5 NR-PCQA methods. The FR-PCQA methods include MSE-p2point \cite{mekuria2016evaluation}, Hausdorff-p2point \cite{mekuria2016evaluation}, MSE-p2plane \cite{tian2017geometric}, Hausdorff-p2plane \cite{tian2017geometric}, PSNR-yuv \cite{torlig2018novel}, PointSSIM \cite{alexiou2020pointssim}, PCQM \cite{meynet2020pcqm}, GraphSIM \cite{yang2020inferring}, MS-GraphSIM \cite{zhang2021ms}, and MPED \cite{yang2022mped}. The NR-PCQA methods include PQA-Net \cite{liu2021pqa}, IT-PCQA \cite{yang2022no}, GPA-Net \cite{shan2023gpa}, ResSCNN \cite{liu2023point}, and MM-PCQA \cite{zhang2022mm}. For a comprehensive comparison, we conduct the experiment in four aspects: 1) We compare the prediction accuracy performance, following the cross-validation configuration in \cref{sec:implementation}. 2) We conduct a visualization analysis for the quality-aware representation learned by CoPA and demonstrate the superiority of our model in terms of distinguishing different distortion patterns. 3) We report the results of cross-dataset evaluation for the NR-PCQA methods to verify the generalization performance of our model. 4) We compare the performance of the NR-PCQA methods with fewer MOS labels for training (fine-tuning for our model).

\noindent\textbf{Comparison of Prediction Accuracy.} The prediction accuracy of all the selected methods on LS-PCQA, SJTU-PCQA and WPC is presented \cref{tab:main}. From the table, we have the following observations: 1) Our model outperforms all the NR-PCQA methods on all three datasets. For example, our model improves the best performance by about $3.20\%$ of SROCC (0.613 vs. 0.594), $1.9\%$ of PLCC (0.636 vs. 0.624), and $6.4\%$ of RMSE (0,161 vs. 0.172) on LS-PCQA. 2) Our method presents competitive performance compared to FR-PCQA methods despite the inaccessibility to the reference information. 3) Our model demonstrates robust performance across the three datasets, regardless of the variations in dataset scale, content, and distortion types.

\begin{figure}[h]
    \centering
    \includegraphics[width=0.47\textwidth]{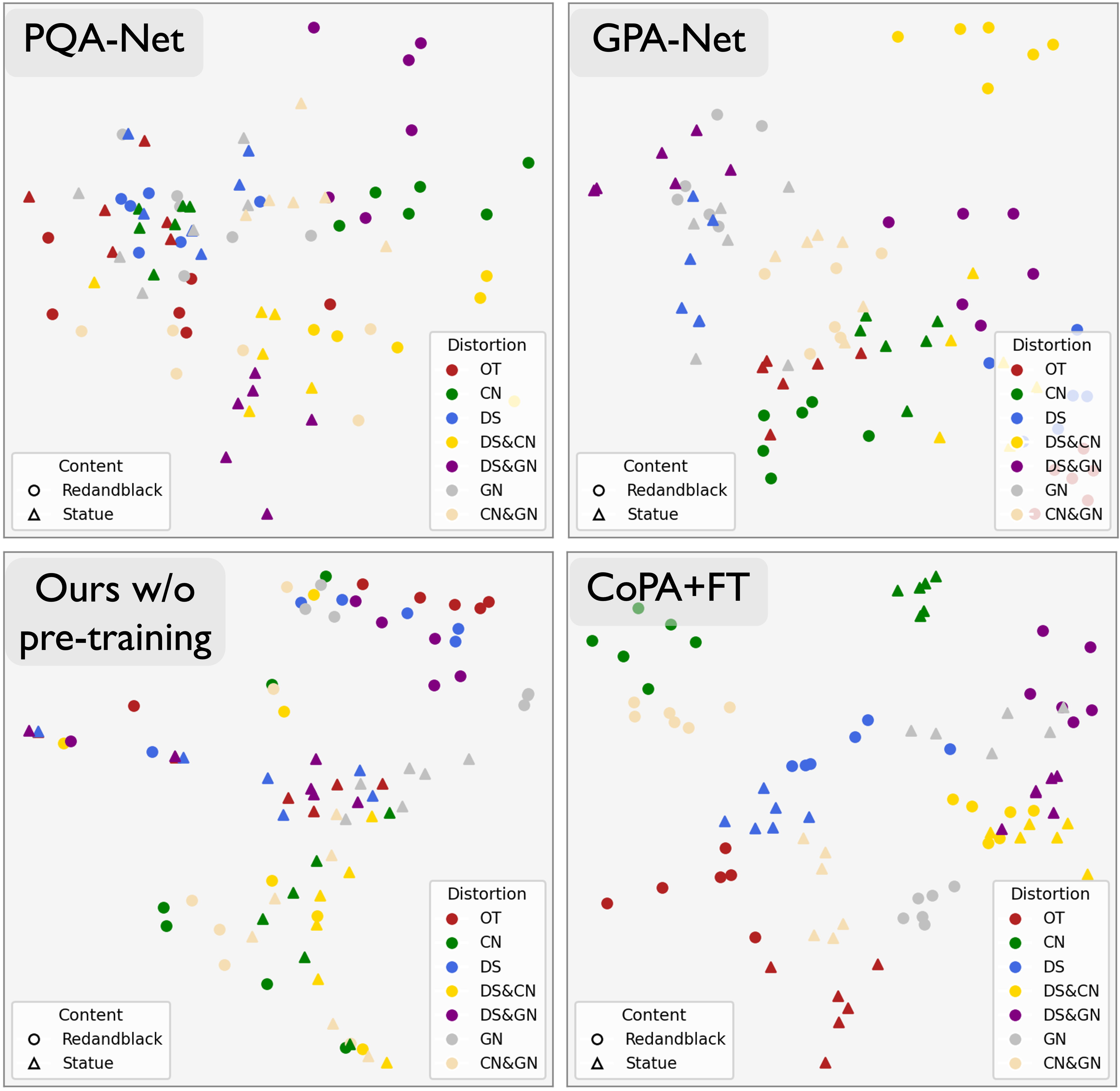}
    \caption{T-SNE embedding of the representation spaces of PQA-Net, GPA-Net, our model without pre-training, and our complete model on testing set of SJTU-PCQA. The scattered points are color and shape encoded according to distortion type and content.}
    \label{fig:visualization}
\end{figure}

\begin{table}[t]
\renewcommand\tabcolsep{3.3pt}
	\centering
 \vspace{-0.3cm}
  \caption{Cross-dataset evaluation for NR-PCQA methods. Training and testing are both conducted on complete datasets. Results of PLCC are reported.}
  \scriptsize
	\begin{tabular}{cc|ccccc}
		\toprule  
		Train & Test & PQA-Net & GPA-Net & ResSCNN & MM-PCQA & CoPA+FT \\ 
		\midrule  
	LS & SJTU & 0.342 & 0.556 & 0.546 & \bf\textcolor{blue}{0.581} & \bf\textcolor{red}{0.644}\\
	LS & WPC & 0.266 & 0.433 & \bf\textcolor{blue}{0.466} & 0.454 &\bf\textcolor{red}{0.516} \\
	WPC & SJTU & 0.235 & 0.553 & 0.572 & \bf\textcolor{blue}{0.612} & \bf\textcolor{red}{0.643} \\
	SJTU & WPC & 0.220 & \bf\textcolor{blue}{0.418} & 0.269 & 0.409 & \bf\textcolor{red}{0.533} \\
		\bottomrule  
	\end{tabular}
	\label{tab:cross}
 \vspace{-0.3cm}
\end{table}

\noindent\textbf{Visualization Analysis.} We visualize the learned representations of PQA-Net, GPA-Net, our model without pre-training and the complete model (\textit{i.e.,} CoPA + fine-tuning) on a unified testing set of a certain fold in SJTU-PCQA. Specifically, we use t-SNE algorithm \cite{van2008visualizing} to embed the representations into 2D feature space, and the visualization results are illustrated in \cref{fig:visualization}. We can see that: 1) Compared to PQA-Net and GPA-Net, our model achieves better clustering results for different distortions, from which we can conclude that the proposed method is able to learn more distortion-discriminative representations than distortion type prediction task. 2) Compared to the proposed network without pre-training, the complete model apparently achieves better discrimination performance, which demonstrates the effectiveness of contrastive pre-training. 

\noindent\textbf{Cross-Dataset Evaluation.} The cross-dataset evaluation is conducted to test the generalizability of the NR-PCQA methods when encountering various data distribution. In \cref{tab:cross}, we mainly train the compared models on the complete LS-PCQA and test the trained model on the complete SJTU-PCQA and WPC, and the result with minimal training loss is recorded. The procedure is repeated for the mutual cross-dataset evaluation between SJTU-PCQA and WPC. Notably, considering that LS-PCQA shares some reference point clouds with SJTU-PCQA, we remove these groups of point clouds from LS-PCQA to avoid information leakage.  From \cref{tab:cross}, we can see that the performance of cross-dataset evaluation is relatively low due to the huge variation with respect to both distortion types and contents. However, our method still outperforms the NR-PCQA methods by a large margin, demonstrating the superior generalizability of CoPA.

\begin{figure}[t]
    \centering
  \includegraphics[width=0.34\textwidth]{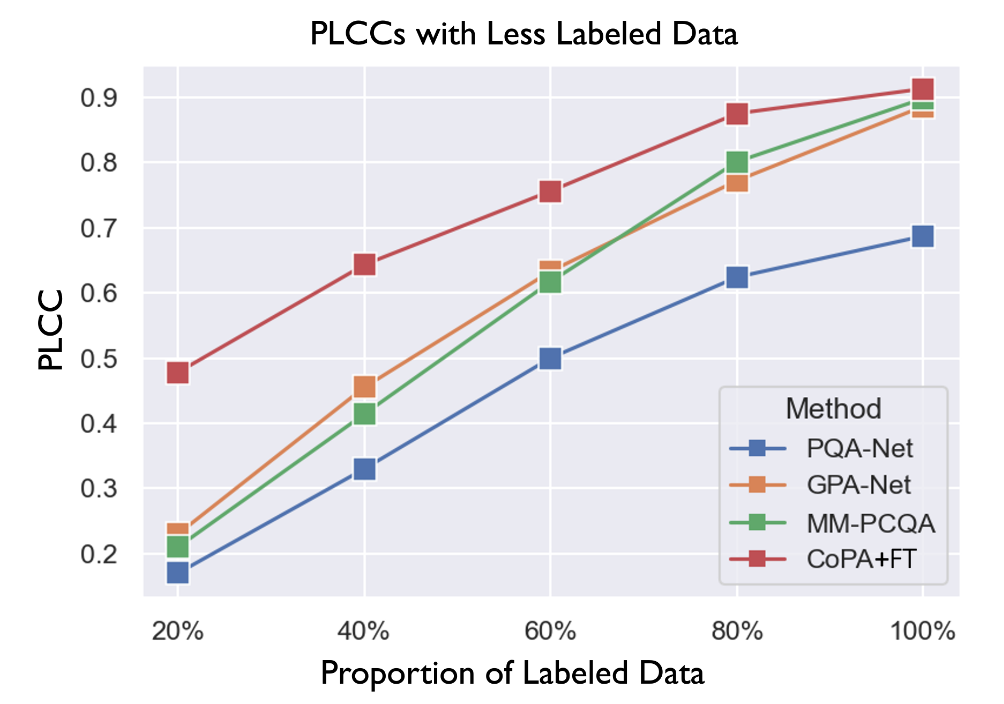}
    \caption{PLCCs of the NR-PCQA methods with less labeled data on SJTU-PCQA. Our pre-trained model outperforms the compared methods by a large margin when the training data is limited.}
    \label{fig:few_shot}
\end{figure}

\noindent\textbf{Performance with Less Labeled Data.} To test the generalization performance of the NR-PCQA models with less labeled data, we compare their performances with different proportions of labeled data. The result is illustrated in \cref{fig:few_shot}. We can see that our model outperforms the compared NR-PCQA methods, and the margin is larger with less training data. Therefore, we can conclude that the pre-training framework can significantly reduce the reliance on annotated data, thus demonstrating its potential to address the issue of label scarcity for the NR-PCQA task.

\subsection{Ablation Studies}
To study the effectiveness of our proposed method, we further investigate the individual contribution of different components and then analyze the effects of the pre-training framework on other NR-PCQA models.

\begin{table}[t]
\renewcommand\arraystretch{1.0}
    \centering
    \caption{Ablation on components of our model on SJTU-PCQA. `\ding{51}' or `\ding{55}' means the setting is preserved or discarded. `Dis.' and `Con.' indicates distortion-wise and content-wise contrast.}
    \label{tab:ablation_components}
    \setlength{\tabcolsep}{2.5pt}
    \scriptsize
    \begin{tabular}{c|ccc|cc}
    \toprule
    Index & Pre-training & Multi-view fusion \tnote{1} & Fine-tuning loss \tnote{1} & SROCC & PLCC\\ \midrule
    \circled{1}&\ding{51} & \ding{51} & \ding{51} & \bf\textcolor{red}{0.897} & \bf\textcolor{red}{0.913} \\ \hdashline
    \circled{2}&\ding{55} & \ding{51} & \ding{51} & 0.806 & 0.893 \\
    \circled{3}& Dis. only & \ding{51} & \ding{51} &   0.856 & 0.887 \\
    \circled{4}& Con. only & \ding{51} & \ding{51}  & \bf\textcolor{blue}{0.889}  & 0.901 \\ \hdashline
    \circled{5} & \ding{51} & Max Pooling  & \ding{51} & 0.841 & 0.872 \\
    \circled{6} & \ding{51} & Average Pooling & \ding{51} & 0.810 & 0.847 \\ \hdashline
    \circled{7} & \ding{51}  & \ding{51} & MSE only & \bf\textcolor{blue}{0.889} & \bf\textcolor{blue}{0.910} \\ 
    \bottomrule
    \end{tabular}
\end{table}

\begin{table}[t]
\renewcommand\arraystretch{0.9}
\renewcommand\tabcolsep{3.3pt}
	\centering
  \caption{PLCCs of using CoPA to pre-train NR-PCQA methods.}
  \scriptsize
	\begin{tabular}{c|cccc}
		\toprule  
		Dataset & PQA-Net & PQA-Net+CoPA & MM-PCQA & MM-PCQA+CoPA \\ 
		\midrule  
	SJTU & 0.687 & 0.749 & 0.898 & 0.910 \\
 WPC & 0.579 & 0.911 & 0.774 & 0.778 \\
		\bottomrule  
	\end{tabular}
	\label{tab:ablation_backbone}
\end{table}

\noindent\textbf{Ablation on Different Components.} In \cref{tab:ablation_components}, we report the results on SJTU-PCQA with the conditions of dropping some components. From \cref{tab:ablation_components}, we have the following oberservations: 1) Seeing \circled{1} and \circled{2} - \circled{4}, the pre-training framework brings the most significant improvements when considering both distortion-wise and content-wise contrast. 2) Seeing \circled{1} and \circled{5} - \circled{6}, the semantic-guided multi-view fusion performs better than the symmetric fusion 
strategies (\textit{i.e.} max and average pooling). 3) Seeing \circled{1} and \circled{7}, the performance is close, which demonstrates the robustness of our model using different fine-tuning loss functions.

\noindent\textbf{Contrastive pre-training on other NR-PCQA Methods.} In \cref{tab:ablation_backbone}, we employ the contrastive pre-training framework in \cref{sec:contrastive} to optimize other projection-based NR-PCQA models (some details like feature dimensions are accordingly modified). We can see that the pre-training framework can fit different backbones and bring noticable gain for existing projection-based NR-PCQA methods.

\section{Conclusion}
In this paper, we propose a novel no-reference point cloud quality method based on contrastive learning. Utilizing local patch mixing to generate anchors without impairing the distortion patterns, the proposed CoPA learns quality-aware representations through both content-wise and distortion-wise contrasts. Moreover, in the fine-tuning stage, we use the semantic-guided multi-view fusion to attentively integrate the quality-aware features from different perspectives. Experiments show that our model presents competitive and generalizable performance compared to the state-of-the-art NR-PCQA methods.
{
    \small
    \bibliographystyle{ieeenat_fullname}
    \bibliography{main}

\begin{thebibliography}{44}
\providecommand{\natexlab}[1]{#1}
\providecommand{\url}[1]{\texttt{#1}}
\expandafter\ifx\csname urlstyle\endcsname\relax
  \providecommand{\doi}[1]{doi: #1}\else
  \providecommand{\doi}{doi: \begingroup \urlstyle{rm}\Url}\fi

\bibitem[Afham et~al.(2022)Afham, Dissanayake, Dissanayake, Dharmasiri,
  Thilakarathna, and Rodrigo]{afham2022crosspoint}
Mohamed Afham, Isuru Dissanayake, Dinithi Dissanayake, Amaya Dharmasiri,
  Kanchana Thilakarathna, and Ranga Rodrigo.
\newblock Crosspoint: Self-supervised cross-modal contrastive learning for 3d
  point cloud understanding.
\newblock In \emph{Proceedings of the IEEE/CVF Conference on Computer Vision
  and Pattern Recognition}, pages 9902--9912, 2022.

\bibitem[Alexiou and Ebrahimi(2020)]{alexiou2020pointssim}
Evangelos Alexiou and Touradj Ebrahimi.
\newblock Towards a point cloud structural similarity metric.
\newblock In \emph{ICMEW}, pages 1--6, 2020.

\bibitem[Chen et~al.(2021)Chen, Li, Wu, Dong, and Shi]{chen2021contrastive}
Pengfei Chen, Leida Li, Jinjian Wu, Weisheng Dong, and Guangming Shi.
\newblock Contrastive self-supervised pre-training for video quality
  assessment.
\newblock \emph{IEEE Transactions on Image Processing}, 31:\penalty0 458--471,
  2021.

\bibitem[Chen et~al.(2020)Chen, Kornblith, Norouzi, and Hinton]{chen2020simple}
Ting Chen, Simon Kornblith, Mohammad Norouzi, and Geoffrey Hinton.
\newblock A simple framework for contrastive learning of visual
  representations.
\newblock In \emph{International conference on machine learning}, pages
  1597--1607. PMLR, 2020.

\bibitem[Deng et~al.(2009)Deng, Dong, Socher, Li, Li, and
  Fei-Fei]{deng2009imagenet}
Jia Deng, Wei Dong, Richard Socher, Li-Jia Li, Kai Li, and Li Fei-Fei.
\newblock Imagenet: A large-scale hierarchical image database.
\newblock In \emph{2009 IEEE Conference on Computer Vision and Pattern
  Recognition}, pages 248--255. Ieee, 2009.

\bibitem[Dosovitskiy et~al.(2020)Dosovitskiy, Beyer, Kolesnikov, Weissenborn,
  Zhai, Unterthiner, Dehghani, Minderer, Heigold, Gelly,
  et~al.]{dosovitskiy2020image}
Alexey Dosovitskiy, Lucas Beyer, Alexander Kolesnikov, Dirk Weissenborn,
  Xiaohua Zhai, Thomas Unterthiner, Mostafa Dehghani, Matthias Minderer, Georg
  Heigold, Sylvain Gelly, et~al.
\newblock An image is worth 16x16 words: Transformers for image recognition at
  scale.
\newblock \emph{arXiv preprint arXiv:2010.11929}, 2020.

\bibitem[Fan et~al.(2022)Fan, Zhang, Sun, Min, Liu, Zhou, He, Wang, and
  Zhai]{fan2022no}
Yu Fan, Zicheng Zhang, Wei Sun, Xiongkuo Min, Ning Liu, Quan Zhou, Jun He,
  Qiyuan Wang, and Guangtao Zhai.
\newblock A no-reference quality assessment metric for point cloud based on
  captured video sequences.
\newblock In \emph{2022 IEEE 24th International Workshop on Multimedia Signal
  Processing (MMSP)}, pages 1--5. IEEE, 2022.

\bibitem[He et~al.(2016)He, Zhang, Ren, and Sun]{he2016deep}
Kaiming He, Xiangyu Zhang, Shaoqing Ren, and Jian Sun.
\newblock Deep residual learning for image recognition.
\newblock In \emph{Proceedings of the IEEE conference on computer vision and
  pattern recognition}, pages 770--778, 2016.

\bibitem[He et~al.(2020)He, Fan, Wu, Xie, and Girshick]{he2020momentum}
Kaiming He, Haoqi Fan, Yuxin Wu, Saining Xie, and Ross Girshick.
\newblock Momentum contrast for unsupervised visual representation learning.
\newblock In \emph{Proceedings of the IEEE/CVF conference on computer vision
  and pattern recognition}, pages 9729--9738, 2020.

\bibitem[Kingma and Ba(2014)]{kingma2014adam}
Diederik~P Kingma and Jimmy Ba.
\newblock Adam: A method for stochastic optimization.
\newblock \emph{arXiv preprint arXiv:1412.6980}, 2014.

\bibitem[Liu et~al.(2021{\natexlab{a}})Liu, Yuan, Su, Liu, Wang, Yang, and
  Hou]{liu2021pqa}
Qi Liu, Hui Yuan, Honglei Su, Hao Liu, Yu Wang, Huan Yang, and Junhui Hou.
\newblock Pqa-net: Deep no reference point cloud quality assessment via
  multi-view projection.
\newblock \emph{IEEE Transactions on Circuits and Systems for Video
  Technology}, 31\penalty0 (12):\penalty0 4645--4660, 2021{\natexlab{a}}.

\bibitem[Liu et~al.(2022)Liu, Su, Duanmu, Liu, and Wang]{liu2022perceptual}
Qi Liu, Honglei Su, Zhengfang Duanmu, Wentao Liu, and Zhou Wang.
\newblock Perceptual quality assessment of colored 3d point clouds.
\newblock \emph{IEEE Transactions on Visualization and Computer Graphics},
  2022.

\bibitem[Liu et~al.(2023)Liu, Yang, Xu, and Yang]{liu2023point}
Yipeng Liu, Qi Yang, Yiling Xu, and Le Yang.
\newblock Point cloud quality assessment: Dataset construction and
  learning-based no-reference metric.
\newblock \emph{ACM Transactions on Multimedia Computing, Communications and
  Applications}, 19\penalty0 (2s):\penalty0 1--26, 2023.

\bibitem[Liu et~al.(2021{\natexlab{b}})Liu, Lin, Cao, Hu, Wei, Zhang, Lin, and
  Guo]{liu2021swin}
Ze Liu, Yutong Lin, Yue Cao, Han Hu, Yixuan Wei, Zheng Zhang, Stephen Lin, and
  Baining Guo.
\newblock Swin transformer: Hierarchical vision transformer using shifted
  windows.
\newblock In \emph{Proceedings of the IEEE/CVF international conference on
  computer vision}, pages 10012--10022, 2021{\natexlab{b}}.

\bibitem[Madhusudana et~al.(2022)Madhusudana, Birkbeck, Wang, Adsumilli, and
  Bovik]{madhusudana2022image}
Pavan~C Madhusudana, Neil Birkbeck, Yilin Wang, Balu Adsumilli, and Alan~C
  Bovik.
\newblock Image quality assessment using contrastive learning.
\newblock \emph{IEEE Transactions on Image Processing}, 31:\penalty0
  4149--4161, 2022.

\bibitem[Mekuria et~al.(2016)Mekuria, Li, Tulvan, and
  Chou]{mekuria2016evaluation}
R Mekuria, Z Li, C Tulvan, and P Chou.
\newblock Evaluation criteria for point cloud compression.
\newblock \emph{ISO/IEC MPEG}, \penalty0 (16332), 2016.

\bibitem[Meynet et~al.(2020)Meynet, Nehm{\'e}, Digne, and
  Lavou{\'e}]{meynet2020pcqm}
Gabriel Meynet, Yana Nehm{\'e}, Julie Digne, and Guillaume Lavou{\'e}.
\newblock Pcqm: A full-reference quality metric for colored 3d point clouds.
\newblock In \emph{QoMEX}, pages 1--6, 2020.

\bibitem[Mu et~al.(2023)Mu, Shao, Chai, Liu, Chen, and Jiang]{mu2023multi}
Baoyang Mu, Feng Shao, Xiongli Chai, Qiang Liu, Hangwei Chen, and Qiuping
  Jiang.
\newblock Multi-view aggregation transformer for no-reference point cloud
  quality assessment.
\newblock \emph{Displays}, 78:\penalty0 102450, 2023.

\bibitem[Oord et~al.(2018)Oord, Li, and Vinyals]{oord2018representation}
Aaron van~den Oord, Yazhe Li, and Oriol Vinyals.
\newblock Representation learning with contrastive predictive coding.
\newblock \emph{arXiv preprint arXiv:1807.03748}, 2018.

\bibitem[Paszke et~al.(2019)Paszke, Gross, Massa, Lerer, Bradbury, Chanan,
  Killeen, Lin, Gimelshein, Antiga, et~al.]{paszke2019pytorch}
Adam Paszke, Sam Gross, Francisco Massa, Adam Lerer, James Bradbury, Gregory
  Chanan, Trevor Killeen, Zeming Lin, Natalia Gimelshein, Luca Antiga, et~al.
\newblock Pytorch: An imperative style, high-performance deep learning library.
\newblock \emph{Advances in neural information processing systems}, 32, 2019.

\bibitem[Qi et~al.(2017)Qi, Su, Mo, and Guibas]{qi2017pointnet}
Charles~R Qi, Hao Su, Kaichun Mo, and Leonidas~J Guibas.
\newblock Pointnet: Deep learning on point sets for 3d classification and
  segmentation.
\newblock In \emph{Proceedings of the IEEE conference on computer vision and
  pattern recognition}, pages 652--660, 2017.

\bibitem[Ravi et~al.(2020)Ravi, Reizenstein, Novotny, Gordon, Lo, Johnson, and
  Gkioxari]{ravi2020accelerating}
Nikhila Ravi, Jeremy Reizenstein, David Novotny, Taylor Gordon, Wan-Yen Lo,
  Justin Johnson, and Georgia Gkioxari.
\newblock Accelerating 3d deep learning with pytorch3d.
\newblock \emph{arXiv preprint arXiv:2007.08501}, 2020.

\bibitem[Saha et~al.(2023)Saha, Mishra, and Bovik]{saha2023re}
Avinab Saha, Sandeep Mishra, and Alan~C Bovik.
\newblock Re-iqa: Unsupervised learning for image quality assessment in the
  wild.
\newblock In \emph{Proceedings of the IEEE/CVF Conference on Computer Vision
  and Pattern Recognition}, pages 5846--5855, 2023.

\bibitem[Shan et~al.(2023)Shan, Yang, Ye, Zhang, Xu, Xu, and Liu]{shan2023gpa}
Ziyu Shan, Qi Yang, Rui Ye, Yujie Zhang, Yiling Xu, Xiaozhong Xu, and Shan Liu.
\newblock Gpa-net: No-reference point cloud quality assessment with multi-task
  graph convolutional network.
\newblock \emph{IEEE Transactions on Visualization and Computer Graphics},
  2023.

\bibitem[Sun et~al.(2022)Sun, Min, Lu, and Zhai]{sun2022deep}
Wei Sun, Xiongkuo Min, Wei Lu, and Guangtao Zhai.
\newblock A deep learning based no-reference quality assessment model for ugc
  videos.
\newblock In \emph{Proceedings of the 30th ACM International Conference on
  Multimedia}, pages 856--865, 2022.

\bibitem[Sutskever et~al.(2013)Sutskever, Martens, Dahl, and
  Hinton]{sutskever2013importance}
Ilya Sutskever, James Martens, George Dahl, and Geoffrey Hinton.
\newblock On the importance of initialization and momentum in deep learning.
\newblock In \emph{International conference on machine learning}, pages
  1139--1147. PMLR, 2013.

\bibitem[Tao et~al.(2021)Tao, Jiang, Jiang, and Yu]{tao2021point}
Wen-xu Tao, Gang-yi Jiang, Zhi-di Jiang, and Mei Yu.
\newblock Point cloud projection and multi-scale feature fusion network based
  blind quality assessment for colored point clouds.
\newblock In \emph{Proceedings of the 29th ACM International Conference on
  Multimedia}, pages 5266--5272, 2021.

\bibitem[Tian et~al.(2017)Tian, Ochimizu, Feng, Cohen, and
  Vetro]{tian2017geometric}
Dong Tian, Hideaki Ochimizu, Chen Feng, Robert Cohen, and Anthony Vetro.
\newblock Geometric distortion metrics for point cloud compression.
\newblock In \emph{IEEE ICIP}, pages 3460--3464, 2017.

\bibitem[Tliba et~al.(2022)Tliba, Chetouani, Valenzise, and
  Dufaux]{tliba2022representation}
Marouane Tliba, Aladine Chetouani, Giuseppe Valenzise, and Fr{\'e}deric Dufaux.
\newblock Representation learning optimization for 3d point cloud quality
  assessment without reference.
\newblock In \emph{2022 IEEE International Conference on Image Processing
  (ICIP)}, pages 3702--3706. IEEE, 2022.

\bibitem[Tliba et~al.(2023)Tliba, Chetouani, Valenzise, and
  Dufaux]{tliba2023pcqa}
Marouane Tliba, Aladine Chetouani, Giuseppe Valenzise, and Fr{\'e}deric Dufaux.
\newblock Pcqa-graphpoint: Efficient deep-based graph metric for point cloud
  quality assessment.
\newblock In \emph{ICASSP 2023-2023 IEEE International Conference on Acoustics,
  Speech and Signal Processing (ICASSP)}, pages 1--5. IEEE, 2023.

\bibitem[Torlig et~al.(2018)Torlig, Alexiou, Fonseca, de~Queiroz, and
  Ebrahimi]{torlig2018novel}
Eric~M Torlig, Evangelos Alexiou, Tiago~A Fonseca, Ricardo~L de Queiroz, and
  Touradj Ebrahimi.
\newblock A novel methodology for quality assessment of voxelized point clouds.
\newblock In \emph{Applications of Digital Image Processing XLI}, pages
  174--190, 2018.

\bibitem[Van~der Maaten and Hinton(2008)]{van2008visualizing}
Laurens Van~der Maaten and Geoffrey Hinton.
\newblock Visualizing data using t-sne.
\newblock \emph{Journal of machine learning research}, 9\penalty0 (11), 2008.

\bibitem[Xiong et~al.(2023)Xiong, Wu, Luo, Suo, and Gao]{xiong2023psi}
Jian Xiong, Sifan Wu, Wang Luo, Jinli Suo, and Hao Gao.
\newblock $\psi$-net: Point structural information network for no-reference
  point cloud quality assessment.
\newblock In \emph{ICASSP 2023-2023 IEEE International Conference on Acoustics,
  Speech and Signal Processing (ICASSP)}, pages 1--5. IEEE, 2023.

\bibitem[Yang et~al.(2020{\natexlab{a}})Yang, Chen, Ma, Xu, Tang, and
  Sun]{yang2020predicting}
Qi Yang, Hao Chen, Zhan Ma, Yiling Xu, Rongjun Tang, and Jun Sun.
\newblock Predicting the perceptual quality of point cloud: A 3d-to-2d
  projection-based exploration.
\newblock \emph{IEEE Transactions on Multimedia}, 23:\penalty0 3877--3891,
  2020{\natexlab{a}}.

\bibitem[Yang et~al.(2020{\natexlab{b}})Yang, Ma, Xu, Li, and
  Sun]{yang2020inferring}
Qi Yang, Zhan Ma, Yiling Xu, Zhu Li, and Jun Sun.
\newblock Inferring point cloud quality via graph similarity.
\newblock \emph{IEEE transactions on pattern analysis and machine
  intelligence}, 44\penalty0 (6):\penalty0 3015--3029, 2020{\natexlab{b}}.

\bibitem[Yang et~al.(2022{\natexlab{a}})Yang, Liu, Chen, Xu, and
  Sun]{yang2022no}
Qi Yang, Yipeng Liu, Siheng Chen, Yiling Xu, and Jun Sun.
\newblock No-reference point cloud quality assessment via domain adaptation.
\newblock In \emph{Proceedings of the IEEE/CVF Conference on Computer Vision
  and Pattern Recognition}, pages 21179--21188, 2022{\natexlab{a}}.

\bibitem[Yang et~al.(2022{\natexlab{b}})Yang, Zhang, Chen, Xu, Sun, and
  Ma]{yang2022mped}
Qi Yang, Yujie Zhang, Siheng Chen, Yiling Xu, Jun Sun, and Zhan Ma.
\newblock Mped: Quantifying point cloud distortion based on multiscale
  potential energy discrepancy.
\newblock \emph{IEEE Transactions on Pattern Analysis and Machine
  Intelligence}, 45\penalty0 (5):\penalty0 6037--6054, 2022{\natexlab{b}}.

\bibitem[Ye et~al.(2023)Ye, Ni, Xu, Wang, Chen, and Eldar]{ye2023fedfm}
Rui Ye, Zhenyang Ni, Chenxin Xu, Jianyu Wang, Siheng Chen, and Yonina~C Eldar.
\newblock Fedfm: Anchor-based feature matching for data heterogeneity in
  federated learning.
\newblock \emph{IEEE Transactions on Signal Processing}, 2023.

\bibitem[Zhang et~al.(2022{\natexlab{a}})Zhang, Guo, Zhang, Li, Miao, Cui,
  Qiao, Gao, and Li]{zhang2022pointclip}
Renrui Zhang, Ziyu Guo, Wei Zhang, Kunchang Li, Xupeng Miao, Bin Cui, Yu Qiao,
  Peng Gao, and Hongsheng Li.
\newblock Pointclip: Point cloud understanding by clip.
\newblock In \emph{Proceedings of the IEEE/CVF Conference on Computer Vision
  and Pattern Recognition}, pages 8552--8562, 2022{\natexlab{a}}.

\bibitem[Zhang et~al.(2021)Zhang, Yang, and Xu]{zhang2021ms}
Yujie Zhang, Qi Yang, and Yiling Xu.
\newblock Ms-graphsim: Inferring point cloud quality via multiscale graph
  similarity.
\newblock In \emph{Proceedings of the 29th ACM International Conference on
  Multimedia}, pages 1230--1238, 2021.

\bibitem[Zhang et~al.(2022{\natexlab{b}})Zhang, Sun, Min, Wang, Lu, and
  Zhai]{zhang2022no}
Zicheng Zhang, Wei Sun, Xiongkuo Min, Tao Wang, Wei Lu, and Guangtao Zhai.
\newblock No-reference quality assessment for 3d colored point cloud and mesh
  models.
\newblock \emph{IEEE Transactions on Circuits and Systems for Video
  Technology}, 32\penalty0 (11):\penalty0 7618--7631, 2022{\natexlab{b}}.

\bibitem[Zhang et~al.(2022{\natexlab{c}})Zhang, Sun, Min, Zhou, He, Wang, and
  Zhai]{zhang2022mm}
Zicheng Zhang, Wei Sun, Xiongkuo Min, Quan Zhou, Jun He, Qiyuan Wang, and
  Guangtao Zhai.
\newblock Mm-pcqa: Multi-modal learning for no-reference point cloud quality
  assessment.
\newblock \emph{arXiv preprint arXiv:2209.00244}, 2022{\natexlab{c}}.

\bibitem[Zhao et~al.(2023)Zhao, Yuan, Sun, Li, and Wen]{zhao2023quality}
Kai Zhao, Kun Yuan, Ming Sun, Mading Li, and Xing Wen.
\newblock Quality-aware pre-trained models for blind image quality assessment.
\newblock In \emph{Proceedings of the IEEE/CVF Conference on Computer Vision
  and Pattern Recognition}, pages 22302--22313, 2023.

\bibitem[Zhou et~al.(2022)Zhou, Yang, Jiang, Zhai, and Lin]{zhou2022blind}
Wei Zhou, Qi Yang, Qiuping Jiang, Guangtao Zhai, and Weisi Lin.
\newblock Blind quality assessment of 3d dense point clouds with structure
  guided resampling.
\newblock \emph{arXiv preprint arXiv:2208.14603}, 2022.

\end{thebibliography}
}


\end{document}